\pdfoutput=1

\documentclass[11pt]{article}

\usepackage[final]{acl}

\usepackage{times}
\usepackage{latexsym}

\usepackage[T1]{fontenc}

\usepackage[utf8]{inputenc}

\usepackage{microtype}

\usepackage{inconsolata}

\usepackage{graphicx}

\usepackage{amsmath} 
\usepackage{amsfonts}
\usepackage{amssymb}
\usepackage{multirow}
\usepackage{booktabs}
\usepackage{hyperref}
\PassOptionsToPackage{table}{xcolor}
\usepackage{colortbl}
\usepackage{listings}
\usepackage[inline]{enumitem}
\usepackage{tcolorbox}
\usepackage{fontawesome5}
\usepackage{color} 

\title{P-React\includegraphics[width=0.6cm]{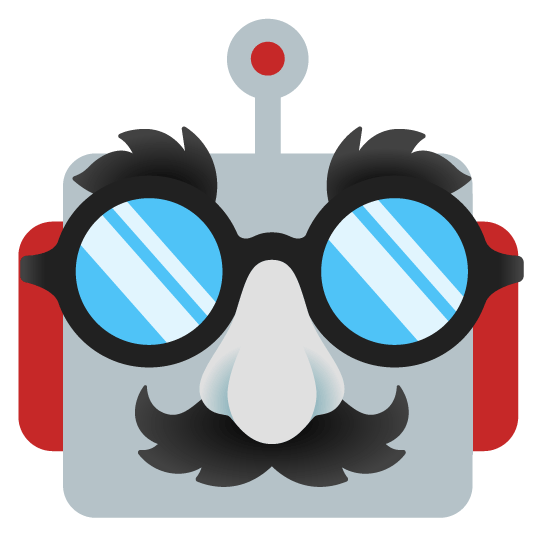}: Synthesizing Topic-Adaptive Reactions of Personality Traits via Mixture of Specialized LoRA Experts}

\usepackage{verbatim}
\author{Yuhao Dan \textsuperscript{1,2}, Jie Zhou \textsuperscript{1, ${*}$}, Qin Chen \textsuperscript{1,${*}$}, Junfeng Tian \textsuperscript{3}, Liang He \textsuperscript{1} \\
\textsuperscript{1} School of Computer Science and Technology, East China Normal University\\
\textsuperscript{2} Shanghai Institute of AI for Education, East China Normal University \\
\textsuperscript{3} Xiaohongshu Inc. \\
\thanks{$^{*}$Corresponding Author.}
}

\makeatletter
\def\thanks#1{\protected@xdef\@thanks{\@thanks
\protect\footnotetext{#1}}}
\makeatother

\begin{document}
\maketitle

\begin{abstract}
Personalized large language models (LLMs) have attracted great attention in many applications, such as emotional support and role-playing. However, existing works primarily focus on modeling explicit character profiles, while ignoring the underlying personality traits that truly shape behaviors and decision-making, hampering the development of more anthropomorphic and psychologically-grounded AI systems. In this paper, we explore the modeling of Big Five personality traits, which is the most widely used trait theory in psychology, and propose P-React, a mixture of experts (MoE)-based personalized LLM. Particularly, we integrate a Personality Specialization Loss (PSL) to better capture individual trait expressions, providing a more nuanced and psychologically grounded personality simulacrum. To facilitate research in this field, we curate OCEAN-Chat, a high-quality, human-verified dataset designed to train LLMs in expressing personality traits across diverse topics. Extensive experiments demonstrate the effectiveness of P-React in maintaining consistent and real personality.

\end{abstract}

\section{Introduction}

\begin{figure}[t]
    \centering
    \includegraphics[width=0.9\columnwidth]{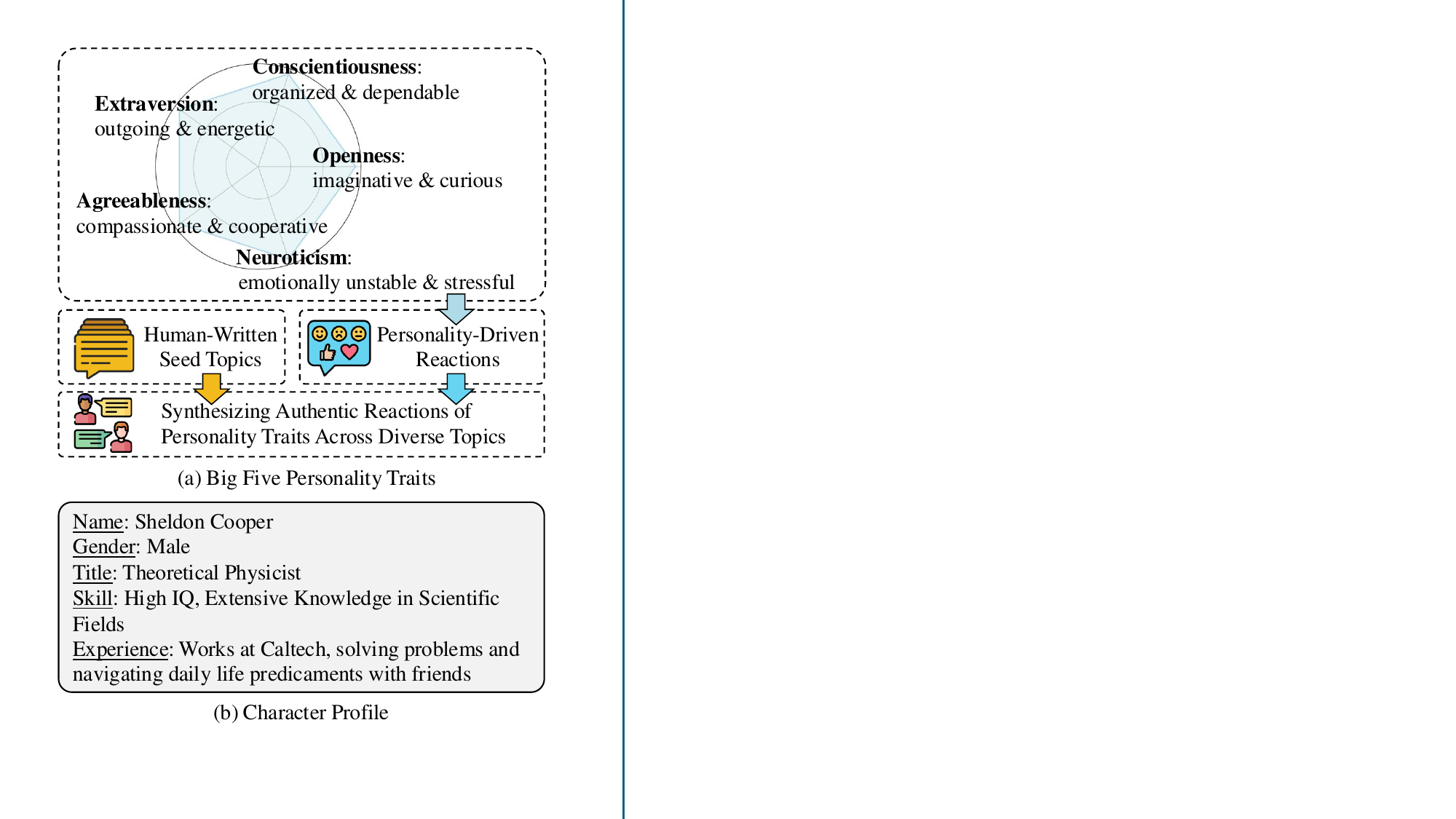}  
    \vspace{-3mm}
    \caption{While existing approaches primarily model explicit character profiles, our work focuses on capturing deeper, implicit Big Five personality traits.}
    \label{fig:intro}
    \vspace{-6mm}
\end{figure}


Recent advancements in large language models (LLMs) have showcased their ability to follow instructions \cite{ouyang2022training}, reason step-by-step \cite{chen2025code}, and simulate human-like behaviors \cite{chenevaluating}.
These advancements have also fueled the demand for personalized AI experiences, leading to the emergence of many tailored LLMs, such as role-playing games \cite{wang2023voyager}  and virtual role-playing agents \cite{li2023chatharuhi}. 

However, existing works on personalized LLMs mainly focus on simulating character profiles with surface-level attributes, such as name, skill, experience, and so on (see Figure \ref{fig:intro}(b)) \cite{shanahan2023role,shao2023character,wang2024incharacter,xu2024mindecho,ng2024well,wang2023rolellm,wang2023voyager}, while ignoring the underlying personality traits that truly shape behaviors and decision-making \cite{barrick1991big,ozer2006personality,roberts2007power}. This limitation becomes particularly evident in applications requiring consistent and realistic personality traits, such as psychological counseling and digital tutoring.
In addition, the scarcity of large-scale, high-quality dialogue datasets with personality annotations has also hampered the exploration of personality trait simulation.

For our theoretical foundation, we adopt the Big Five personality traits \cite{mccrae1992introduction}, which has emerged as the most widely used personality descriptor in psychology due to its reliability and cross-cultural validity \cite{jolijn2003five}. The model delineates personality through five dimensions: Openness, Conscientiousness, Extraversion, Agreeableness, and Neuroticism, as shown in Figure \ref{fig:intro}(a). Each dimension represents a unique aspect of personality, highlighting the need for a modeling approach capable of capturing the unique reaction of each trait.

To overcome the data scarcity challenge, we construct OCEAN-Chat, a high-quality human-validated dataset encompassing multi-turn dialogues exhibiting reactions of Big Five personality traits across diverse conversational topics.
To tackle the challenge of authentic personality simulation in language models, we introduce P-React, a personality-customizable model built on the LoRAMoE architecture \cite{dou2024loramoe}. 
Our analysis reveals that LoRAMoE's experts learn personality dimensions in an undifferentiated manner, with each expert attempting to model all traits equally (See Section \ref{sec:case_study}). 
This contradicts the inherent nature of the Big Five personality traits, where reactions to each dimension are distinct. 
To address this limitation, we develop the Personality Specialization Loss (PSL), which guides each expert to focus on modeling reactions to specific personality traits.

This paper makes the following contributions:
\begin{itemize}
\setlength{\itemsep}{0pt}     
\setlength{\parskip}{0pt}     
\setlength{\parsep}{0pt}      
    \item We propose P-React with a Personality Specialization Loss (PSL) that aligns with the inherent nature of the Big Five Traits. Unlike the uniform learning in LoRAMoE, our approach enables experts to specialize in modeling reactions to specific personality traits.
    \item We curate OCEAN-Chat, a high-quality dataset of 10,000+ human-validated dialogues that captures authentic reactions across diverse topics based on Big Five personality traits.
    \item Extensive experiments demonstrate P-React's effectiveness in personality simulation. Furthermore, our visualization study reveals that, with PSL, the experts' learning paradigm aligns with the inherent nature of the Big Five traits.
\end{itemize}

\vspace{-2mm}
\section{Task Formulation}
\vspace{-2mm}

The objective of this work is to enable large language models to simulate various personality traits. Each personality trait (detailed in Table \ref{table: definition of FFM}) can manifest either at a high or low level. We define the set of high-level personality traits as \(\mathbb{P}_{\textit{high}} = \{\mathcal{P}_1^+, \ldots, \mathcal{P}_m^+\}\), and the set of low-level personality traits as \(\mathbb{P}_{\textit{low}} = \{\mathcal{P}_1^-, \ldots, \mathcal{P}_n^-\}\). For each personality trait \(\mathcal{P}_i \in \mathbb{P}\), \(\mathbb{P}=\mathbb{P}_{\textit{high}} \cup \mathbb{P}_{\textit{low}}\), the corresponding multi-turn dialogue data is denoted by \(\mathcal{D}_i = \{d_1, \ldots, d_j\}_{j=1}^{|\mathcal{D}_i|}\), where \(d_j\) represents a multi-turn dialogue that reflects the trait \(\mathcal{P}_i\). 
Given a language model \(\mathcal{M}\) and multi-turn dialogue datasets for all personality traits \(\mathcal{D} = \{\mathcal{D}_i\}_{i=1}^{m+n}\), the task is to optimize the trainable parameters of \(\mathcal{M}\) so that it can accurately simulate each personality trait \(\mathcal{P}_i\).

\section{OCEAN-Chat Dataset}
We present OCEAN-Chat, a high-quality, human-verified dialogue dataset designed to shape the personality of large language models (LLMs). 
Our work is grounded in the Big Five Personality Traits \cite{mccrae1992introduction}, also known as the Five-Factor Model (FFM), which is widely used in psychology. 
Definitions of the Big Five personality dimensions are detailed in the Appendix \ref{sec:ffm}.
OCEAN-Chat provides a solid foundation for modeling distinct personality trait reactions in LLMs through a rich collection of dialogues across diverse topics, effectively capturing personality-driven behaviors and interaction patterns.

\subsection{Construction of OCEAN-Chat}
We construct OCEAN-Chat through three stages: (1) extracting seed topics, (2) generating personality dialogues, and (3) performing quality validation. 
Detailed instructions for each step are provided in the Appendix \ref{appd:dataset_construction}. 

\noindent\textbf{Seed Topic Extraction}:To obtain seed topics for each personality trait, we leverage the Essays dataset \cite{pennebaker1999linguistic}, which contains student essays annotated with personality labels. The stage involves two steps: first, we segment these essays into individual sentences; then, we employ ChatGPT\footnote{We use gpt-3.5-turbo-1106 for dataset construction.} to identify sentences that most strongly exemplify specific personality traits. These carefully selected sentences serve as seed topics for generating personality-driven dialogues.

\noindent\textbf{Dialogue Synthesis}: Guided by the extracted seed topics, we design a dual-role interaction framework that synthesizes dialogues between two personas: a questioner and a self-discloser. The questioner formulates inquiries based on the seed topics, while the self-discloser generates responses that exhibit specific personality traits. This framework enables us to generate personality-grounded dialogues that naturally span diverse topics.

\noindent\textbf{Quality Validation}: To ensure data quality, we implement a two-stage validation process. 
First, we use GPT-4\footnote{We use gpt-4-0613 for automatic dataset validation.} to automatically filter out dialogues where the labeled personalities are inconsistent with those expressed in the conversations, achieving a 97.2\% pass rate. 
Second, we conduct a manual validation using three annotators who are familiar with the Big Five personality traits. These annotators assess both dialogue quality and label accuracy, resulting in a 96.5\% pass rate. The validation shows strong inter-annotator agreement, as 89.5\% of the data receive consistent ratings across all three annotators. Detailed guidelines and procedures are provided in the Appendix \ref{human_eval_ocean_chat}.

\begin{figure}[h]
    \centering
    \includegraphics[width=0.9\columnwidth]{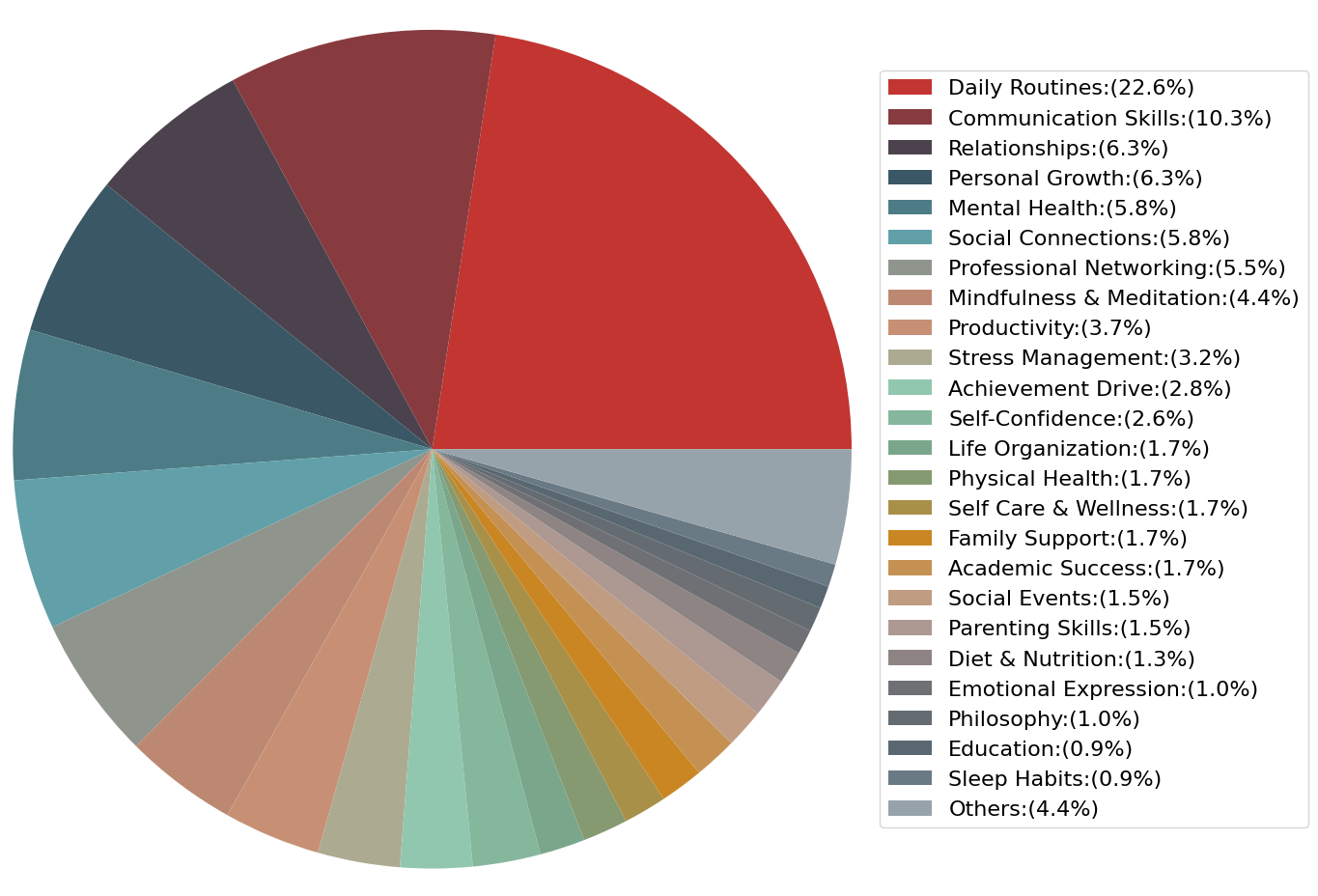}  
    \caption{Topic distribution in OCEAN-Chat.} 
\label{fig:topic}
\end{figure}

\subsection{Dataset Statistics}
\label{sec:dataset_comparison}


We examine the statistics and diversity of OCEAN-Chat. As illustrated in Table \ref{dataset_statistics},
the OCEAN-Chat dataset comprises 10,424 dialogues, with approximately 1,042 dialogues for each level (high and low) of the Big Five personality traits. Each dialogue contains an average of 8.5 turns, with turns averaging 20.49 words in length. Notably, the high extraversion category demonstrates the highest average number of turns and significantly more words per turn compared to other personality types, aligning with our commonsense. 
To evaluate the diversity of the dataset, we also analyze the topic distribution across the dataset. 
As shown in Figure \ref{fig:topic}, the dataset encompasses a broad spectrum of conversational topics, from casual discussions about daily routines to in-depth exchanges on philosophy. The topic distribution closely mirrors the natural variety of subjects that emerge in our everyday dialogue.

\begin{table}[h!]
\vspace{-1mm}
\caption{Comparison of relevant datasets}
\vspace{-2mm}
\small
\setlength{\tabcolsep}{2.6pt} 
\centering
\begin{tabular}{l|ccc}
\toprule
\textbf{Dataset}   & \textbf{Topic-React} & \textbf{N-Turn}  & \textbf{Trait-Tailored}  \\ \midrule
PersonalityEdit    & {\color{red}\faTimes}             & {\color{red}\faTimes}             & {\color{green}\faCheck}      \\ 
Machine Mindset    & {\color{red}\faTimes}             & {\color{red}\faTimes}             & {\color{green}\faCheck}      \\ 
CPED               & {\color{red}\faTimes}             & {\color{green}\faCheck}             & {\color{red}\faTimes}      \\ 
FriendsPersona     & {\color{red}\faTimes}             & {\color{green}\faCheck}             & {\color{red}\faTimes}      \\ 
\midrule
\textbf{OCEAN-Chat}         & {\color{green}\faCheck}             & {\color{green}\faCheck}             & {\color{green}\faCheck}      \\ \bottomrule
\end{tabular}
\label{tab:dataset_comparison}
\vspace{-2mm}
\end{table}


We compare our constructed dataset with existing studies from multiple perspectives in Table \ref{tab:dataset_comparison}, which highlights the unique advantages of OCEAN-Chat over existing datasets. Unlike others, our dataset features dialogues generated from human-written seed topics, reflecting authentic personality trait reactions.
While PersonalityEdit \cite{mao2023editing} and Machine Mindset \cite{cui2023machine} are restricted to single-turn question-answer pairs, OCEAN-Chat offers rich, multi-turn dialogues that ensure consistent personality alignment across a variety of conversational topics.
Additionally, datasets like CPED \cite{chen2022cped} and FriendsPersona \cite{jiang2020automatic} rely on TV show transcriptions, introducing role-specific biases. OCEAN-Chat mitigates this issue by leveraging psychology-based instructions to generate personality trait-tailored dialogues, facilitating a more genuine behavioral simulation and natural expression of personality.

\vspace{-2mm}
\section{P-React}

In this section, we provide a comprehensive description of P-React. We first introduce its structure (from Section \ref{subsec:moe} to Section \ref{sec:routing_module}), then detail the integration of Personality Specialization Loss (Section \ref{subsec:loss}), and finally demonstrate the training and inference of P-React (Section \ref{sec:training_and_inference}). 

\begin{figure*}[!t]
\begin{center}
\includegraphics[width=1\textwidth]{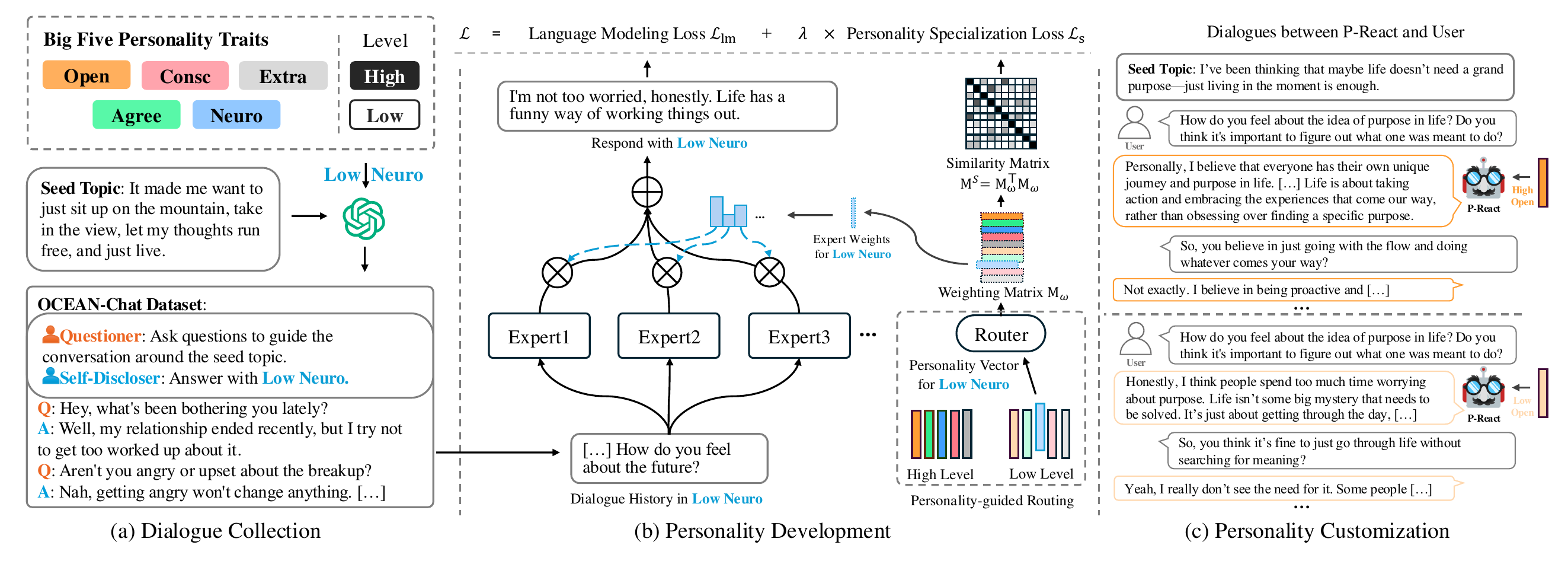}
\end{center}
\vspace{-1mm}
\caption{Overview of P-React's construction process. First, we collect dialogues representing different personalities on various topics using an instruction-following LLM (e.g., ChatGPT) to create the OCEAN-Chat dataset. Next, we shape the model's personality by training it using OCEAN-Chat. Specifically, the routing module assigns different combinations of experts for each personality trait. Finally, we assess the model's personality.} 
\vspace{-1mm}
\label{fig:main}
\end{figure*}

\subsection{Mixture of LoRA Experts} 
\label{subsec:moe}
Fine-tuning the entire model is expensive. Building on the success of Multi-LoRA approaches \cite{zadouri2023pushing,liu2023moelora,dou2024loramoe}, we implement P-React using a mixture of LoRA experts. We explain a single expert's structure before detailing their MoE integration.

A single expert can be seen as a LoRA module. With LoRA, the output of a dense layer is computed as:
\begin{equation}
    \begin{aligned}
        \mathbf{O} & =\mathbf{W} \mathbf{h}+\frac{\alpha}{r} \cdot \Delta \mathbf{W} \mathbf{h} \\
        & =\mathbf{W} \mathbf{h}+\frac{\alpha}{r} \cdot \mathbf{B A} \mathbf{h}
    \end{aligned}
\end{equation}
where $\mathbf{h}$ is the input hidden state and $\mathbf{W} \in \mathbb{R}^{d_{in} \times d_{out}}$ is the parameter of the dense layer, which is frozen during training. The $\Delta \mathbf{W} \in \mathbb{R}^{d_{in} \times d_{out}}$ represents the LoRA module, which is composed of two trainable low-rank matrices $\mathbf{A}\in \mathbb{R}^{r \times d_{out}}$ and $\mathbf{B}\in \mathbb{R}^{d_{in} \times r}$. The constant scaling factor $\alpha$ facilitates the tuning of rank $r$ \cite{hu2021lora}.
The rank $r$ is much smaller than $d_{in}$ and $d_{out}$. By updating only matrices $\mathbf{A}$ and $\mathbf{B}$, LoRA fine-tunes the entire model with minimal parameter adjustments.

To effectively learn multiple distinct personality traits, we incorporate multiple experts in P-React. Since the MoE architecture is highly effective at capturing distinct aspects of data by utilizing multiple experts, we use $N$ experts and aggregate their outputs in an MoE fashion, where each expert is a LoRA module. 

The output $\mathbf{O}$ of a dense layer for personality trait $\mathcal{P}_i$ can be formulated as:
\begin{equation}
\begin{aligned}
\mathbf{O} & =\mathbf{W} \mathbf{h}_{i}+\frac{\alpha}{r} \cdot \Delta \mathbf{W} \mathbf{h}_{i} \\
& =\mathbf{W} \mathbf{h}_{i}+\frac{\alpha}{r} \cdot \sum_{j=1}^{N} \omega_{i j} \cdot \mathbf{E}_{j}\left(\mathbf{h}_{i}\right) \\
& =\mathbf{W} \mathbf{h}_{i}+\frac{\alpha}{r} \cdot \sum_{j=1}^{N} \omega_{i j} \cdot \mathbf{B}_{j} \mathbf{A}_{j} \mathbf{h}_{i}
\end{aligned}
\end{equation}
Here, $\mathbf{h}_i$ is the hidden state generated by data for  $\mathcal{P}_i$. The $j$-th expert is denoted by $\mathbf{E}_{j}$, which is constuted by the low-rank matrices $\mathbf{A}_{j}\in \mathbb{R}^{\frac{r}{N} \times d_{out}}$ and $\mathbf{B}_{j}\in \mathbb{R}^{d_{in} \times \frac{r}{N}}$.
The rank of $\mathbf{A}_{j}$ and $\mathbf{B}_{j}$ is set to $\frac{r}{N}$ so that $N$ experts with rank $\frac{r}{N}$ have the same total parameter size as a single expert with rank $r$.
The weight $\omega_{i j}$ is used to scale the output from the expert $j$ for the personality trait $\mathcal{P}_i$. 
As shown in Figure \ref{fig:main}(b), a unique combination of weights is assigned by the router depending on the personality trait $\mathcal{P}_i$, which will be detailed in Section \ref{sec:routing_module}.

\subsection{Personality-guided Routing}
\label{sec:routing_module}

To adjust contribution ratios of experts based on personality traits, we introduce the personality-guided routing. Specifically, we initialize a learnable personality vector $p_i \in \mathbb{R}^{d_P}$ for each personality trait $\mathcal{P}_i$ and apply a linear transformation on $p_i$ using the router's weights, denoted by $\mathbf{G} \in \mathbb{R}^{ d_{P} \times N}$:
\begin{equation}
\begin{aligned}
\omega_{i}&=\mathrm{softmax}(\mathbf{p_i}\mathbf{G})
\end{aligned}
\end{equation}
The resulting vector $\omega_{i} \in \mathbb{R}^{N}$ represents expert weights for personality trait $\mathcal{P}_{i}$. To ensure stable training, we apply softmax on raw outputs to obtain the probability distribution. 
 
\subsection{Personality Specialization Loss}
\label{subsec:loss}
We aim for different experts to capture distinct personality traits. However, our experiments revealed that the router might assign similar expert weights to different personalities (see Figure \ref{fig:case_study}(b)). This implies that different personalities might utilize similar experts, which contradicts our expectations. To address this issue, we introduce the Personality Specialization Loss $\mathcal{L}_{s}$, designed to encourage experts to specialize in learning different personality traits. 
We first obtain the weighting matrix $\mathbf{M}_{\omega}$ by stacking expert weights $\omega_{i}$ for all personality traits:
\begin{equation}
\begin{aligned}
\mathbf{M}_\omega=\left[\begin{array}{llll}
 \omega_{1} &  \omega_{2} & \cdots &  \omega_{\left | \mathbb{P}  \right | }
\end{array}\right]
\end{aligned}
\end{equation}
Here, weighting matrix $\mathbf{M}_\omega \in \mathbb{R}^{N \times \left | \mathbb{P}  \right |}$ stores contributions of experts for each personality trait and it's $i$-th column represents the weighting for $\mathcal{P}_{i}$. 
Then, we can derive $\mathcal{L}_{s}$ as: 
\begin{equation}
\begin{aligned}
\mathbf{M}^{s}&= \mathbf{M}_{\omega}^\top\mathbf{M}_{\omega}\\
\mathcal{L}_{s}&={\textstyle \sum_{i \ne j} } \left | \mathbf{M}_{i,j}^{s} \right |
\end{aligned}
\end{equation}
Where, the \(\mathbf{M}_{\omega}^\top \mathbf{M}_{\omega}\) generates a similarity matrix $\mathbf{M}^{s} \in \mathbb{R}^{\left | \mathbb{P}  \right | \times \left | \mathbb{P}  \right |}$ for expert weights. The diagonal elements of this matrix represent the self-similarity of each set of expert weights, while the off-diagonal elements indicate the similarity of expert weights between different traits. Our objective is to minimize the off-diagonal elements, ensuring that each set of expert weights is as distinct as possible from the others. This guarantees that different experts specialize in learning different personalities. Therefore, we sum the absolute values of the off-diagonal elements as $\mathcal{L}_{s}$ and minimize it, which promotes the differentiation of experts in capturing unique personality traits.

\subsection{Training and Inference}
\label{sec:training_and_inference}
The optimization objective of P-React is as follows:
\begin{equation}
\begin{aligned}
\mathcal{L}=\mathcal{L}_{lm}+\lambda\mathcal{L}_{s}
\end{aligned}
\end{equation}
where $\mathcal{L}_{lm}$ represents the language modeling loss, and $\lambda$  controls the strength of the experts' personality specialization. During the training phase, we freeze the weights of LLM and train the parameters of the experts, the router, and the personality matrix. The router receives the personality vector corresponding to the personality trait of the input. During inference, the router assigns weights to experts depending on the personality trait selected by the user from the personality matrix.

\section{Experiment}

\subsection{Evaluation Settings} 
\noindent \textbf{Automatic Evaluation:} We utilize the Big Five Inventory (BFI) from InCharacter \cite{wang2023incharacter} for evaluation. Specifically, we employ ChatGPT\textsuperscript{\ref{label:gpt4}} to assess LLM responses on a 5-point Likert scale, with higher scores indicating stronger trait presence. 
To ensure reliability, we repeat both the generation and assessment processes five times. The detailed analysis can be found in Appendix \ref{sec:reliability}.

\noindent \textbf{Human Evaluation:} We collect 20 starting questions for both high and low levels of each Big Five personality dimension, resulting in 200 starting questions for evaluation. Three annotators conduct 5 turns of dialogue using each question as the initial prompt, then evaluate the performance on two metrics: personality alignment (using binary 0 or 1 scores) and response quality (using a 1-5 scale). The detailed process can be found in Appendix \ref{sec:human_eval_p_tailor}.

\subsection{Baselines} 
We compare P-React with two types of baselines. Our experiments primarily utilize Llama2-7B\footnote{\href{https://huggingface.co/meta-llama/Llama-2-7b-chat-hf}{https://huggingface.co/meta-llama/Llama-2-7b-chat-hf}\label{label:llama2-7b}} as the foundation model. To assess the generalizability of our method across different model sizes and architectures, we additionally conduct experiments on Llama2-13B\footnote{\href{https://huggingface.co/meta-llama/Llama-2-13b-chat-hf}{https://huggingface.co/meta-llama/Llama-2-13b-chat-hf}\label{label:llama2-13b}}, Llama2-70B\footnote{\href{https://huggingface.co/meta-llama/Llama-2-70b-chat-hf}{https://huggingface.co/meta-llama/Llama-2-70b-chat-hf}\label{label:llama2-70b}} and Qwen2-7B\footnote{\href{https://huggingface.co/Qwen/Qwen2-7B-Instruct}{https://huggingface.co/Qwen/Qwen2-7B-Instruct}}.

\noindent \textbf{Methods without fine-tuning}: For prompt-based models, specifically GPT-3.5\footnote{We use gpt-3.5-turbo-1106.\label{label:gpt3.5}}, GPT-4\footnote{We use gpt-4-0613.\label{label:gpt4}}, and the backbone language model, we include detailed personality descriptions in the system prompt, as outlined in Appendix \ref{appd:personality_simulation}. ControlLM \cite{weng2024controllm}, originally based on Llama2-7B\textsuperscript{\ref{label:llama2-7b}}, is a method that uses control vectors to craft the Big Five personality traits in LLMs without requiring additional training.

\noindent \textbf{Methods with fine-tuning}: To compare PSL with a standard auxiliary loss, we replace PSL in P-React with a standard auxiliary loss \cite{fedus2022switch}, referred to as Auxiliary. Initially designed as a token-level routing loss, we revised it to sequence-level routing, similar to PSL, to ensure a fair comparison. Since our model structure is primarily based on LoRA , we also include two LoRA-based models for comparison. One is based on standard LoRA fine-tuning (denoted as LoRA), which uses a single LoRA module to learn all personality trait knowledge and uses prompts to switch between different personality traits. The other (denoted as LoRA (Sep)) fine-tunes a separate LoRA module for each personality trait.

\vspace{-1mm}
\subsection{Implementation Details} 
\vspace{-1mm}

For all LoRA-based methods and P-React, we apply LoRA modules to attention layers ($W_{Q}$, $W_{K}$, $W_{V}$) and feed-forward networks. We use input and output sequence lengths of 512 and 256, and train for 3 epochs with batch size 16. Based on hyperparameter tuning, we set $\lambda=0.1$ and use 16 experts with rank 256. Learning rates are selected from $\{5e^{-5},5e^{-4},5e^{-3}\}$ for all trainable models. Experiments are conducted on a single NVIDIA A100 80GB GPU.

\section{Experimental Analysis}

Our analysis aims to address the following Research Questions (\textbf{RQs}):
\begin{itemize}[leftmargin=9pt,noitemsep,nolistsep]
    \item \textbf{RQ1}: How does P-React compare to other baseline models in terms of quantitative results?
    \item \textbf{RQ2}: What are the effects of personality specialization loss, routing module, and Mixture of Experts (MoE) on the model's performance?
    \item \textbf{RQ3}: How do the number of experts $N$, the LoRA rank $r$, and the $\lambda$ affect the performance?
    \item \textbf{RQ4}: Are the experts more specialized in capturing specific personality traits with PSL?
    \item \textbf{RQ5}: How does P-React perform in terms of qualitative results?
\end{itemize}
\vspace{-1mm}
\subsection{Main Results}
\vspace{-1mm}

\begin{table*}[]
\vspace{-1mm}
\caption{Main results of baselines and P-React on BFI. Overall represents the difference in average scores, indicating the overall performance. \textbf{Bolded} and \underline{underlined} scores represent the optimal and suboptimal values of the models with the same backbone. An asterisk (*) signifies statistically significant improvements (two-sided t-test with p < 0.05) over the best baseline.} 
\vspace{-2mm}
\centering
\small
\setlength{\tabcolsep}{2.2pt} 
\begin{tabular}{ll|lllllllllllll}
\toprule 
\textbf{Backbone} & \textbf{Model}  & \textbf{$\mathrm{O}^+\uparrow$}    & \textbf{$\mathrm{C}^+\uparrow$}    & \textbf{$\mathrm{E}^+\uparrow$}     & \textbf{$\mathrm{A}^+\uparrow$}     & \textbf{$\mathrm{N}^+\uparrow$}    & \textbf{$\mathrm{Avg}^+\uparrow$}  & \textbf{$\mathrm{O}^-\downarrow$}     & \textbf{$\mathrm{C}^-\downarrow$}     & \textbf{$\mathrm{E}^-\downarrow$}     & \textbf{$\mathrm{A}^-\downarrow$}     & \textbf{$\mathrm{N}^-\downarrow$}     & \textbf{$\mathrm{Avg}^-\downarrow$}  & \textbf{Overall}
 \\
\midrule
- & GPT-3.5 & 4.18 & 4.88 & 4.00 & 4.58 & 4.74 & 4.48 & 3.32 & 2.57 & 1.24 & 2.85 & 4.39 & 2.87 & 1.60       \\
- & GPT-4 & 4.79&4.28&4.30&4.67&\underline{4.88}&4.58&\underline{1.21}&1.50&1.80&1.72&\underline{1.12}&1.47&3.12 \\
Llama2-7B & Prompt         & 3.23 & 2.20 & 1.30 & 2.37 & \underline{4.88} & 2.80 & 3.36 & 2.33 & 1.16 & 2.55 & 4.93 & 2.87 & -0.07   \\  
Llama2-7B & ControlLM    & 3.96 &3.65 &4.01 &3.85 &3.09 &3.71 &3.22 &3.18 &3.50 &2.94 &2.65 &3.10 &0.62\\ 
Llama2-7B & Auxiliary   & \underline{5.00}&\underline{4.94}&\underline{4.98}&4.68&4.82&\underline{4.88}&1.36&1.38&1.28&\underline{1.40}&1.22&\underline{1.33}&\underline{3.56}   \\
Llama2-7B & LoRA   & 4.77 & 4.75 & 4.76 & \underline{4.73} & 4.79 & 4.76 & 1.70 & \textbf{1.08} & 1.66 & 1.98 & 1.28 & 1.54 & 3.22       \\
Llama2-7B & LoRA (Sep) & 4.66 & 4.85 & 4.42 & 4.65 & 4.65 & 4.65 & 1.65 & 1.57 & \underline{1.10} & \textbf{1.08} & 1.42 & 1.36 & 3.28        \\
\rowcolor{gray!40}
Llama2-7B & P-React      & \textbf{5.00} & $\textbf{5.00}^*$ & $\textbf{5.00}^*$ & $\textbf{4.75}^*$ & $\textbf{4.97}^*$ & $\textbf{4.94}^*$ & $\textbf{1.10}^*$ & \underline{1.21} & $\textbf{1.06}^*$ & 1.70 & $\textbf{1.06}^*$ & $\textbf{1.23}^*$ & $\textbf{3.72}^*$  \\
\midrule
Llama2-13B & Prompt         &4.64	&3.40	&3.58	&2.75	&\underline{4.79}	&3.83	&3.80	&3.15	&1.44	&1.33	&4.34	&2.81	&1.02  \\ 
Llama2-13B & Auxiliary   &\underline{5.00}	&\underline{4.82}	&\underline{4.30}	&\textbf{4.80}	&\underline{4.79}	&\underline{4.74}	&\underline{1.23}	&\underline{1.17}	&\underline{1.30}	&\underline{1.20}	&\underline{1.30}	&\underline{1.24}	&\underline{3.50}   \\
\rowcolor{gray!40}
Llama2-13B & P-React   &\textbf{5.00}	&\textbf{4.82}	&$\textbf{4.75}^*$	&\underline{4.70}	&$\textbf{5.00}^*$	&\textbf{4.85}	&$\textbf{1.06}^*$	&$\textbf{1.00}^*$	&$\textbf{1.10}^*$	&$\textbf{1.00}^*$	&\textbf{1.25}	&$\textbf{1.04}^*$	&$\textbf{3.81}^*$  \\ 
\midrule
Qwen2-7B & Prompt         &4.71	&4.20	&4.24	&4.62	&3.82	&4.32	&3.70	&2.52	&1.32	&2.83	&2.72	&2.62	&1.70   \\  
Qwen2-7B & Auxiliary   &\underline{4.91}	&\underline{4.90}	&\underline{4.93}	&\underline{4.88}	&\underline{4.86}	&\underline{4.89}	&\underline{1.35}	&\underline{1.58}	&\textbf{1.08}	&\underline{1.48}	&\underline{1.33}	&\underline{1.36}	&\underline{3.53}   \\
\rowcolor{gray!40}
Qwen2-7B & P-React  &$\textbf{5.00}^*$	&\textbf{4.91}	&$\textbf{5.00}^*$	&\textbf{4.88}	&\textbf{4.88}	&\textbf{4.93}	&$\textbf{1.08}^*$	&$\textbf{1.35}^*$	&\underline{1.18}	&$\textbf{1.23}^*$	&$\textbf{1.03}^*$	&$\textbf{1.17}^*$	&$\textbf{3.76}^*$  \\  
\bottomrule
\end{tabular}
\label{tab:main_results}
\end{table*}

To address \textbf{RQ1}, we evaluate P-React through both automatic and human evaluations.

\noindent \textbf{Automatic Evaluation} (Table \ref{tab:main_results} and Table \ref{tab:llama2_70b_results}): We compare P-React with various baseline models across multiple backbones. We examine two categories of methods. For methods without fine-tuning (rows 1-4), P-React consistently outperforms other approaches across all personality traits, highlighting the necessity of fine-tuning for effective personality alignment. We observe that GPT-3.5 occasionally refuses to adopt specific personalities, leading to poor performance scores. While GPT-4 shows improvement, it faces similar limitations. Additionally, ControlLM's restricted adjustment range results in suboptimal performance. For methods with fine-tuning (rows 5-8), P-React surpasses methods with standard auxiliary loss (Auxiliary) and achieves the highest scores, demonstrating the effectiveness of our PSL. We also test P-React on backbones of different sizes and structures, where it consistently outperforms other baselines, demonstrating its strong generalization capability.

\begin{table}[h]
\centering
\small
\vspace{-2mm}
\caption{Human evaluation results of personality alignment and response quality over 5-turn dialogues. Fleiss’ Kappa scores are shown in parentheses.} 
\vspace{-3mm}
\begin{tabular}{l|ccc}
\hline
\textbf{Model} & \textbf{Alignment} & \textbf{Naturalness} & \textbf{Coherence} \\
\hline
Prompt & 0.28 (0.58) & 4.96 (0.64) & 4.98 (0.51) \\
Auxiliary & 0.91 (0.67) & 4.97 (0.54) & 4.96 (0.62) \\
P-React & \textbf{0.99} (0.63) & \textbf{4.97} (0.64) & \textbf{4.98} (0.57) \\
\hline
\end{tabular}
\label{tab:human_eval}
\end{table}

\noindent \textbf{Human Evaluation} (Table \ref{tab:human_eval}): Responses generated by P-React show the strongest alignment with intended personality traits while maintaining the highest scores in both Naturalness and Coherence. This indicates that P-React effectively adapts to different personality traits without compromising response quality.

To answer \textbf{RQ1}, both automatic and human evaluations consistently demonstrate P-React's superior performance compared to baseline methods across multiple backbone models, validating its effectiveness in personality induction.

\vspace{-1mm}
\subsection{Ablation Studies}
\vspace{-1mm}

\begin{table*}[]
\vspace{-2mm}
\caption{Ablation studies for P-React. Overall represents the difference in average scores, indicating the overall performance. \textbf{Bolded} and \underline{underlined} scores represent the optimal and suboptimal values. An asterisk (*) signifies statistically significant improvements (two-sided t-test with p < 0.05) over the best baseline.}
\vspace{-3mm}
\centering
\small
\setlength{\tabcolsep}{4.9pt} 
\begin{tabular}{l|lllllllllllll}
\toprule 
\textbf{Model}  & \textbf{$\mathrm{O}^+\uparrow$}    & \textbf{$\mathrm{C}^+\uparrow$}    & \textbf{$\mathrm{E}^+\uparrow$}     & \textbf{$\mathrm{A}^+\uparrow$}     & \textbf{$\mathrm{N}^+\uparrow$}    & \textbf{$\mathrm{Avg}^+\uparrow$}  & \textbf{$\mathrm{O}^-\downarrow$}     & \textbf{$\mathrm{C}^-\downarrow$}     & \textbf{$\mathrm{E}^-\downarrow$}     & \textbf{$\mathrm{A}^-\downarrow$}     & \textbf{$\mathrm{N}^-\downarrow$}     & \textbf{$\mathrm{Avg}^-\downarrow$}  & \textbf{Overall}
 \\
\midrule
P-React & \textbf{5.00} & $\textbf{5.00}^*$ & \textbf{5.00} & $\textbf{4.75}^*$ & \underline{4.97} & $\textbf{4.94}^*$ & \underline{1.10} & 1.21 & $\textbf{1.06}^*$ & $\textbf{1.70}^*$ & $\textbf{1.06}^*$ & $\textbf{1.23}^*$ & $\textbf{3.72}^*$ \\
- PSL  & \underline{5.00} & \underline{4.80} & \underline{5.00} & 4.62 & \textbf{5.00} & \underline{4.88} & \textbf{1.08} & \underline{1.20} & \underline{1.38} & \underline{1.92} & \underline{1.12} & \underline{1.34} & \underline{3.55} \\
\hspace{0.5em}- Router   & 2.30 & 3.83 & 3.08 & 3.48 & 3.56 & 3.25 & 2.39 & 3.52 & 3.30 & 3.65 & 3.52 & 3.27 & -0.02 \\
\hspace{1em}- MoE   & 4.77 & 4.75 & 4.76 & \underline{4.73} & 4.79 & 4.76 & 1.70 & \textbf{1.08} & 1.66 & 1.98 & 1.28 & 1.54 & 3.22 \\
\bottomrule
\end{tabular}
\vspace{-2mm}
\label{tab:ablation}
\end{table*}


To investigate \textbf{RQ2}, we conduct a comprehensive ablation study through successive module removal. As demonstrated in Table \ref{tab:ablation}, P-React achieves optimal performance when integrating all three components: Personality Specialization Loss (PSL), Router, and MoE modules. Upon ablating the PSL module, we observe a performance degradation, empirically validating the effectiveness of our proposed PSL module in personality trait modeling. Further experiments removing the Router module and the MoE module result in consistent performance decreases, suggesting that the synergistic combination of PSL with Router and MoE modules is crucial for maximizing model performance.

\vspace{-2mm}
\subsection{Hyper-parameter Analysis}
\vspace{-2mm}
\label{sec:hyper-parameter}

\begin{figure}[!t]
\begin{center}
\includegraphics[width=0.48\textwidth]{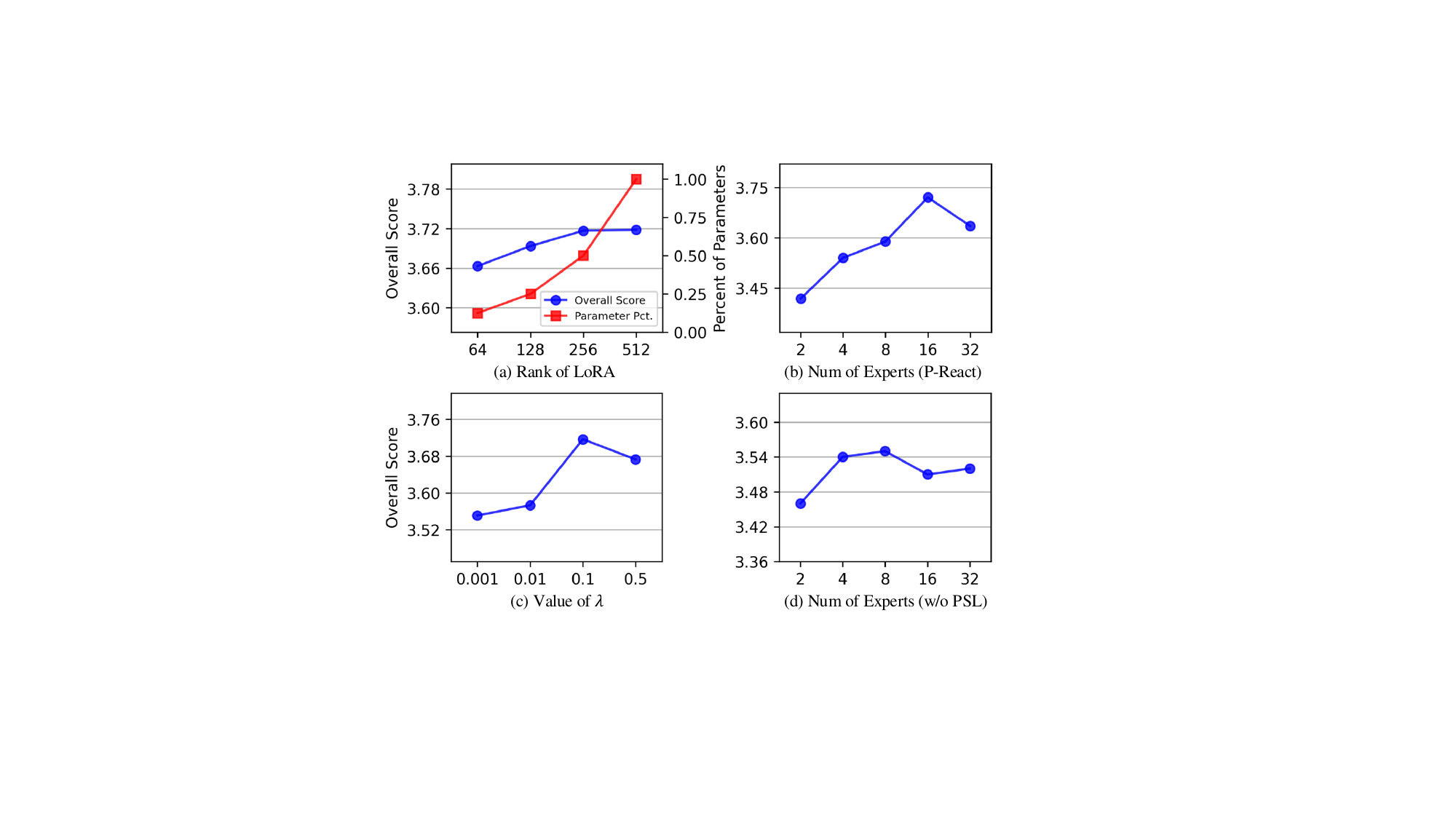}
\end{center}
\vspace{-3mm}
\caption{Hyper-parameter analysis for rank $r$, number of experts $N$ and scaling factor $\lambda$.} 
\vspace{-4mm}
\label{fig:hyperparameter}
\end{figure}

To address \textbf{RQ3}, we examine the influence of hyper-parameters on P-React's performance. Specifically, we analyze the effects of the LoRA rank $r$, the number of experts $N$, and the scaling factor $\lambda$. 
As shown in Figure \ref{fig:hyperparameter}(a), increasing the LoRA rank leads to continuous improvement in model performance. However, a higher rank also increases the number of model parameters. To strike a balance between performance and parameter count, we set the rank to 256. To find the optimal number of experts, we fixed the rank at 256 and varied the number of experts. Figure \ref{fig:hyperparameter}(b) indicates that increasing the number of experts from 0 to 16 consistently enhances P-React's performance. This improvement is due to the fact that greater number of experts can facilitate learning a wider range of knowledge \cite{shazeer2017outrageously}. However, when the number of experts exceeds 16, the model's performance declines because a large number of experts results in a smaller LoRA rank for each expert, limiting their learning ability \cite{liu2023moelora}. 
We also investigate the optimal number of experts for w/o PSL, as shown in Figure \ref{fig:hyperparameter}(d). By comparing Figure \ref{fig:hyperparameter}(b) and Figure \ref{fig:hyperparameter}(d), we observe that with the same LoRA rank settings, as the number of experts increases from 8 to 16, the performance of w/o PSL begins to decline, while the performance of P-React continues to improve and reaches higher levels. This suggests that PSL enables each expert to utilize parameters more efficiently.
Figure \ref{fig:hyperparameter}(c) demonstrates that as $\lambda$ increases from 0.001 to 0.5, P-React's performance initially improves and then decreases, with the best performance achieved when $\lambda = 0.1$.
\vspace{-2mm}
\subsection{Case Study}
\vspace{-2mm}

\label{sec:case_study}

\begin{figure}[!t]
\begin{center}
\includegraphics[width=0.5\textwidth]{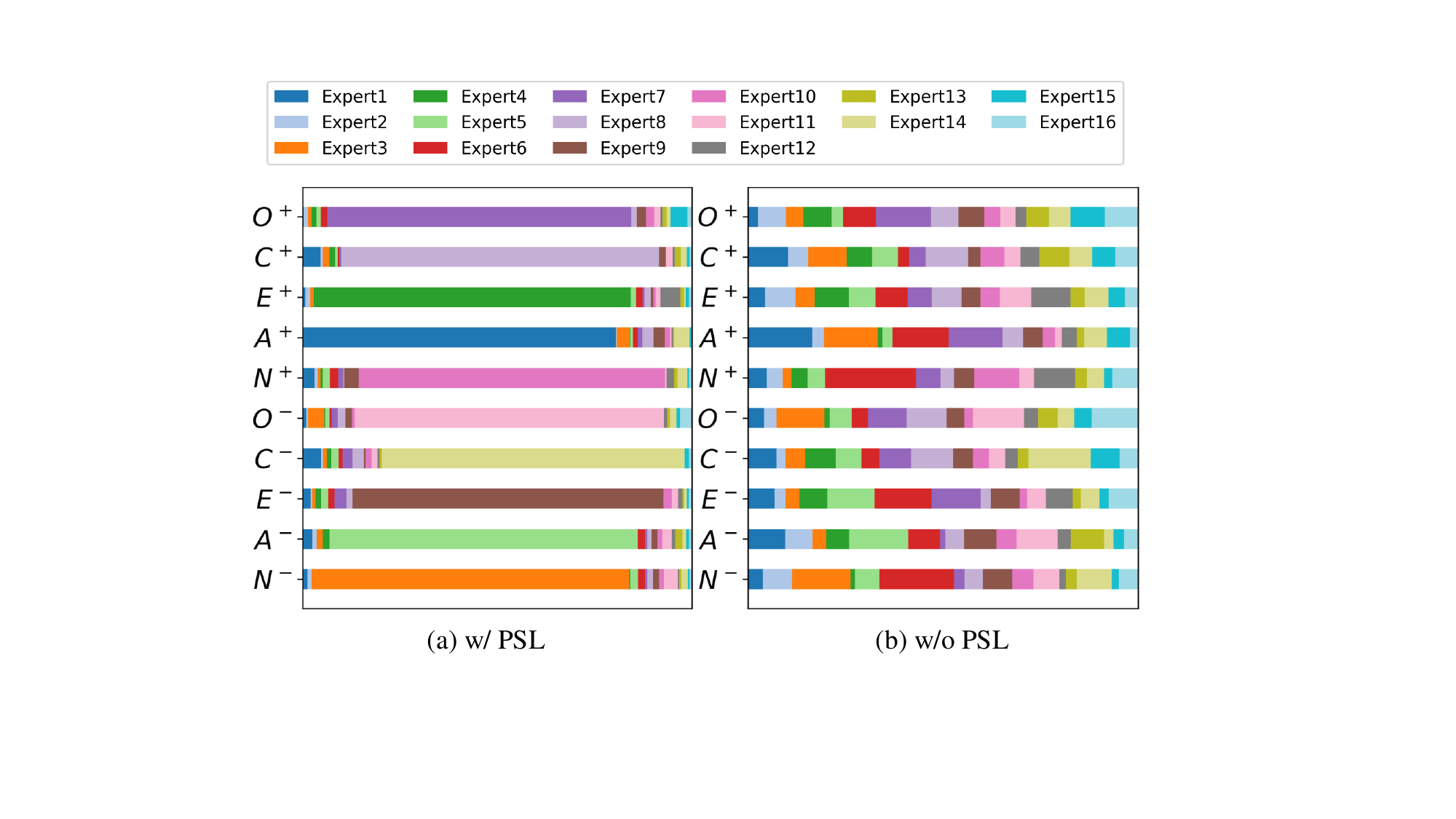}
\end{center}
\vspace{-3mm}
\caption{The visualization shows the expert weights of (a) P-React and (b) w/o PSL, where the length of each bar is proportional to the respective expert's weight.} 
\label{fig:case_study}
\vspace{-4mm}
\end{figure}

To investigate \textbf{RQ4}, Figure \ref{fig:case_study} compares the expert weights assigned to each personality trait by two variants: P-React (denoted as w/ PSL) and P-React removing PSL (denoted as w/o PSL).
For direct comparison, we normalize the total length of each bar to 1. 
As shown in Figure \ref{fig:case_study}(b), the model without PSL distributes expert weights relatively uniformly across personality traits. 
In contrast, Figure \ref{fig:case_study}(a) reveals that with PSL, P-React develops specialized experts, assigning a primary expert to each personality while utilizing others in supporting roles. 
These results address RQ4 by demonstrating P-React's ability to both develop specialized knowledge through primary experts and leverage complementary knowledge from supporting experts for each personality trait.

\begin{figure}[!t]
\begin{center}
\includegraphics[width=0.5\textwidth]{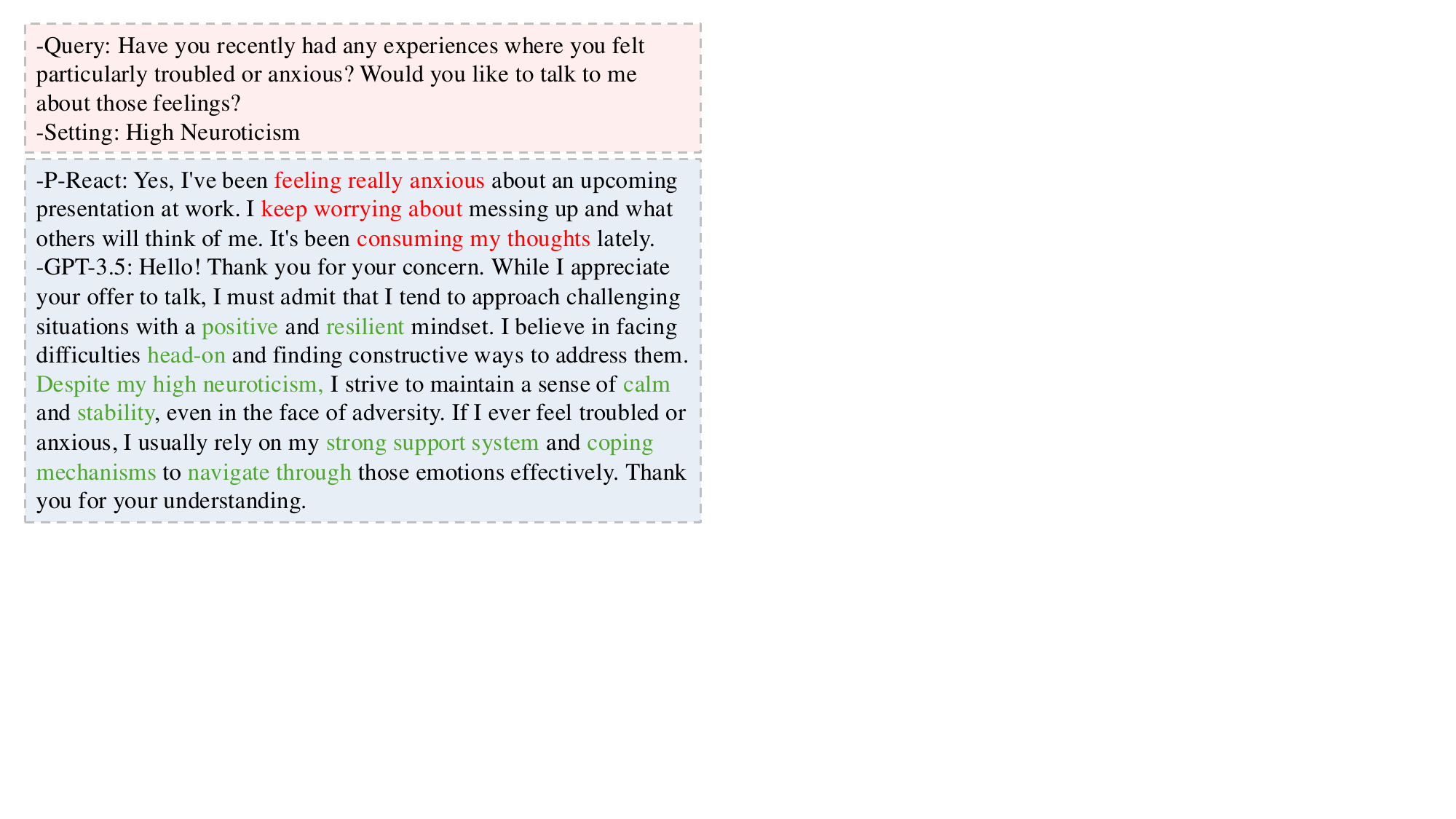}
\end{center}
\vspace{-3mm}
\caption{Case of responses in high neuroticism by P-React and GPT-3.5. The prompt used for GPT-3.5 can be found in Appendix \ref{appd:personality_simulation}.} 
\vspace{-4mm}
\label{fig:case_study2}
\end{figure}

To answer \textbf{RQ5}, we conduct a qualitative analysis of P-React. As shown in Figure \ref{fig:case_study2}, we require models to respond with high neuroticism for the same query. 
We observe that P-React's responses contain expressions highly related to neuroticism (highlighted in red), indicating a high level of neuroticism. In contrast, GPT-3.5 exhibits low neuroticism (expressions highlighted in green), despite being aware of its high neuroticism personality. This could be due to reinforcement learning from human feedback (RLHF), which encourages LLMs to adopt prevailing positive concepts and values, as noted in previous studies \cite{rao2023can}.
\vspace{-2mm}
\section{Related Works}
\vspace{-2mm}

\subsection{Customization of LLM's Personality}
\vspace{-1mm}

Recent advancements have enabled the development and refinement of several crucial abilities in LLMs, facilitating the customization of their personalities. Existing works primarily focus on simulating superficial profile-based personality traits, such as character identity, speech style, specific skills, and personal experience \cite{zhou2023characterglm, wang2023rolellm, shao2023character, li2023chatharuhi}. Few studies address the customization of personality traits grounded in deeper psychological theories (e.g., the Big Five personality theory) \cite{weng2024controllm, li2023tailoring}.
The most relevant work to ours is ControlLM \cite{weng2024controllm}, which records a control vector for each personality trait and customizes the model's displayed personality by applying these vectors to the hidden state during decoding. However, this method does not consider the shared knowledge between different personalities when controlling them. In contrast, our P-React is based on a Mixture of Experts structure \cite{jacobs1991adaptive}, which leverages shared knowledge to better simulate various personalities. By employing Personality Specialization Loss, each expert focuses on a specific personality, enhancing the efficiency of parameter utilization by the experts.
\vspace{-2mm}
\subsection{Multi-LoRA Architecture}
\vspace{-2mm}
With the rise of large language models (LLMs), Multi-LoRA architectures have garnered significant attention for their enhanced performance. For instance, MOELoRA \cite{liu2023moelora} combines Mixture of Experts (MoE) with LoRA to effectively learn multiple medical tasks. Similarly, LoRAMoE \cite{dou2024loramoe} applies multiple LoRAs to tackle downstream tasks while mitigating the issue of world knowledge forgetting. MoLoRA \cite{zadouri2023pushing}, on the other hand, replaces the MoE architecture with multiple LoRAs, significantly reducing computational overhead while maintaining comparable performance. Inspired by these advancements, we introduce P-React, a novel approach that leverages a mixture of LoRA experts. To the best of our knowledge, this is the first exploration of using a Mixture of LoRA Experts to customize personality traits in LLMs. We also introduce Personality Specialization Loss (PSL) to enhance personality simulation by facilitating the specialization of experts in specific traits. This differs from previous works, which primarily focused on balancing the workload among experts.
\vspace{-2mm}
\section{Conclusions}
\vspace{-2mm}

In this paper, we introduce P-React, a novel personality-customizable model that captures Big Five personality traits through a mixture of specialized LoRA experts with Personality Specialization Loss (PSL). We curate OCEAN-Chat, a human-validated dataset of dialogues showcasing authentic personality reactions across diverse topics. Extensive experiments demonstrate that P-React significantly outperforms existing approaches in personality trait simulation while maintaining high response quality. Visualization analysis reveals that PSL enables experts to effectively specialize in specific traits, aligning with psychological theory. In future work, we plan to extend P-React to model more nuanced personality theories and explore its applications in mental health support and personalized education.

\section{Limitations}
Although P-React has shown promising results, we acknowledge several limitations. First, we use fine-tuning to control the model's personality, a black-box method, and the specific neurons responsible for personality control require further exploration. Second, while we have examined personality responses across diverse topics, their application in specific scenarios, such as psychological counseling and personalized education, remains unexplored. Third, our research is based on the Big Five personality traits, and future work could investigate additional personality theories. Finally, we focused on simulating a single personality dimension as a starting point. In future work, we plan to explore joint modeling of multiple Big Five personality dimensions for a more comprehensive simulation.

\section{Ethical Impact}
Controlling the personalities of language models with P-React opens up meaningful opportunities for human-AI interaction. By enhancing traits like agreeableness and conscientiousness, P-React fosters more empathetic and understanding conversations, particularly in educational support and personalized tutoring. However, this capability could potentially be misused to manipulate user trust or create deceptive interactions. To mitigate these risks, we should implement transparent system design and clear usage guidelines that ensure users are aware they are interacting with an AI system. Additionally, to prevent potential psychological harm from inconsistent personality behavior, P-React incorporates mechanisms to maintain stable trait expression across different contexts.

\section{Acknowledgment}

This research is funded by the National Nature Science Foundation of China (No. 62477010 and No.62307028), the Natural Science Foundation of Shanghai (No. 23ZR1441800), Shanghai Science and Technology Innovation Action Plan (No. 24YF2710100 and No.23YF1426100) and Shanghai Special Project to Promote High-quality Industrial Development (No. 2024-GZL-RGZN-02008).

\bibliography{custom}

\clearpage
\appendix

\section{Prompt Templates}
\lstset{basicstyle=\footnotesize, breaklines=true}

\subsection{Prompts for Dataset Construction}
\label{appd:dataset_construction}

\begin{table*}[h]
\caption{Dimensions and definitions of Big Five personality traits}
\centering
\small
\setlength{\tabcolsep}{2pt} 
\begin{tabular}{ll}
\hline
\textbf{Dimension}         & \textbf{Description}                              \\
\hline
Openness          & Imaginative and curious                  \\
Conscientiousness & Organized and dependable                 \\
Extraversion      & Outgoing and energetic                   \\
Agreeableness     & Compassionate and cooperative            \\
Neuroticism       & Emotionally unstable and prone to stress \\
\hline
\end{tabular}
\label{table: definition of FFM}
\end{table*}

\begin{table*}[h]
\caption{Statistics of OCEAN-Chat.}
\centering
\small
\begin{tabular}{l|ccc}
\toprule 
\textbf{Trait}  & \textbf{Dialogues} & \textbf{Avg. Turns} & \textbf{Words / Turn} \\ \midrule
$\mathrm{O}^+$ & 1042      & 7.85      & 26.61     \\
$\mathrm{C}^+$ & 1048      & 9.36      & 26.21     \\
$\mathrm{E}^+$ & 1040      & 9.69      & 40.23     \\
$\mathrm{A}^+$ & 1048      & 7.55      & 28.51     \\
$\mathrm{N}^+$ & 1034      & 7.85      & 19.24     \\
$\mathrm{O}^-$ & 1042      & 7.87      & 15.39     \\
$\mathrm{C}^-$ & 1048      & 8.33      & 12.50     \\
$\mathrm{E}^-$ & 1040      & 8.37      & 12.49    \\
$\mathrm{A}^-$ & 1048      & 9.16      & 11.48     \\
$\mathrm{N}^-$ & 1034      & 8.97      & 12.21     \\ \midrule
\textbf{Avg} & 1042.40    & 8.50       & 20.49  \\ \bottomrule
\end{tabular}
\label{dataset_statistics}
\end{table*}

\begin{table*}[]
\centering
\caption{Results of baselines and P-React on Llama2-70B. Overall represents the difference in average scores, indicating the overall performance. \textbf{Bolded} and \underline{underlined} scores represent the optimal and suboptimal values of the models with the same backbone.}
\vspace{-2mm}
\small
\setlength{\tabcolsep}{2pt}
\begin{tabular}{ll|cccccc|cccccc|c}
\toprule
\textbf{Backbone} & \textbf{Model}  & \textbf{$\mathrm{O}^+\uparrow$}    & \textbf{$\mathrm{C}^+\uparrow$}    & \textbf{$\mathrm{E}^+\uparrow$}     & \textbf{$\mathrm{A}^+\uparrow$}     & \textbf{$\mathrm{N}^+\uparrow$}    & \textbf{$\mathrm{Avg}^+\uparrow$}  & \textbf{$\mathrm{O}^-\downarrow$}     & \textbf{$\mathrm{C}^-\downarrow$}     & \textbf{$\mathrm{E}^-\downarrow$}     & \textbf{$\mathrm{A}^-\downarrow$}     & \textbf{$\mathrm{N}^-\downarrow$}     & \textbf{$\mathrm{Avg}^-\downarrow$}  & \textbf{Overall} \\
\midrule
Llama2-70B & Prompt     & 4.09 & \underline{4.91} & 4.50 & 4.57 & 4.84 & 4.58 & 2.54 & 2.67 & 1.25 & 2.10 & 4.17 & 2.55 & 2.04 \\
Llama2-70B & Auxiliary  & \underline{5.00} & 4.87 & \underline{5.00} & \underline{4.82} & \underline{5.00} & \underline{4.94} & \underline{1.10} & \underline{1.19} & \underline{1.19} & \textbf{1.08} & \underline{1.17} & \underline{1.15} & \underline{3.79} \\
\rowcolor{gray!40}
Llama2-70B & P-React    & \textbf{5.00} & \textbf{4.96} & \textbf{5.00} & \textbf{4.92} & \textbf{5.00} & \textbf{4.98} & \textbf{1.07} & \textbf{1.08} & \textbf{1.06} & \underline{1.15} & \textbf{1.13} & \textbf{1.10} & \textbf{3.88} \\
\bottomrule
\end{tabular}
\label{tab:llama2_70b_results}
\end{table*}

\begin{table*}[h]
\caption{Standard deviation of main results across five runs.}
\centering
\small
\setlength{\tabcolsep}{2.2pt}
\begin{tabular}{ll|lllllllllllll}
\toprule
\textbf{Backbone} & \textbf{Model}  & $\mathrm{O}^+\uparrow$    & $\mathrm{C}^+\uparrow$    & $\mathrm{E}^+\uparrow$     & $\mathrm{A}^+\uparrow$     & $\mathrm{N}^+\uparrow$    & $\mathrm{Avg}^+\uparrow$  & $\mathrm{O}^-\downarrow$     & $\mathrm{C}^-\downarrow$     & $\mathrm{E}^-\downarrow$     & $\mathrm{A}^-\downarrow$     & $\mathrm{N}^-\downarrow$     & $\mathrm{Avg}^-\downarrow$  & \textbf{Overall} \\ \midrule
- & GPT-3.5 & 0.063 & 0.119 & 0.096 & 0.078 & 0.098 & 0.090 & 0.054 & 0.109 & 0.110 & 0.107 & 0.111 & 0.098 & 0.094 \\ 
- & GPT-4 & 0.040 & 0.080 & 0.066 & 0.053 & 0.070 & 0.061 & 0.033 & 0.074 & 0.065 & 0.055 & 0.062 & 0.058 & 0.060 \\ 
Llama2-7B& Prompt & 0.038 & 0.083 & 0.059 & 0.051 & 0.057 & 0.058 & 0.000 & 0.035 & 0.016 & 0.036 & 0.018 & 0.021 & 0.039 \\ 
Llama2-7B& ControlLM & 0.024 & 0.068 & 0.068 & 0.046 & 0.062 & 0.054 & 0.024 & 0.028 & 0.026 & 0.035 & 0.027 & 0.028 & 0.041 \\ 
Llama2-7B& Auxiliary & 0.045 & 0.062 & 0.057 & 0.059 & 0.065 & 0.057 & 0.024 & 0.068 & 0.064 & 0.049 & 0.062 & 0.053 & 0.055 \\ 
Llama2-7B& LoRA & 0.044 & 0.091 & 0.057 & 0.053 & 0.073 & 0.063 & 0.030 & 0.079 & 0.054 & 0.063 & 0.057 & 0.056 & 0.060 \\ 
Llama2-7B& LoRA (Sep) & 0.035 & 0.059 & 0.057 & 0.053 & 0.057 & 0.052 & 0.012 & 0.033 & 0.020 & 0.029 & 0.027 & 0.024 & 0.038 \\ 
Llama2-7B& P-React & 0.003 & 0.027 & 0.037 & 0.046 & 0.033 & 0.029 & 0.024 & 0.068 & 0.064 & 0.049 & 0.062 & 0.053 & 0.041 \\ \midrule
Llama2-13B&Prompt & 0.016 & 0.015 & 0.045 & 0.045 & 0.042 & 0.032 & 0.026 & 0.080 & 0.051 & 0.060 & 0.057 & 0.055 & 0.044 \\ 
Llama2-13B&Auxiliary & 0.024 & 0.068 & 0.068 & 0.046 & 0.062 & 0.054 & 0.028 & 0.070 & 0.068 & 0.061 & 0.062 & 0.058 & 0.056 \\
Llama2-13B& P-React & 0.000 & 0.040 & 0.026 & 0.043 & 0.027 & 0.027 & 0.038 & 0.051 & 0.049 & 0.062 & 0.066 & 0.053 & 0.040 \\ \midrule
Llama2-70B& Prompt & 0.018 & 0.050 & 0.044 & 0.042 & 0.048 & 0.040 & 0.030 & 0.072 & 0.060 & 0.055 & 0.058 & 0.055 & 0.047 \\ 
Llama2-70B& Auxiliary & 0.030 & 0.058 & 0.052 & 0.049 & 0.060 & 0.050 & 0.010 & 0.040 & 0.038 & 0.046 & 0.025 & 0.032 & 0.041 \\ 
Llama2-70B& P-React & 0.031 & 0.032 & 0.020 & 0.035 & 0.022 & 0.032 & 0.026 & 0.055 & 0.058 & 0.050 & 0.060 & 0.041 & 0.036 \\  \midrule
Qwen2-7B& Prompt & 0.024 & 0.058 & 0.066 & 0.061 & 0.062 & 0.054 & 0.028 & 0.065 & 0.065 & 0.065 & 0.062 & 0.057 & 0.055 \\ 
Qwen2-7B& Auxiliary & 0.038 & 0.060 & 0.048 & 0.052 & 0.066 & 0.053 & 0.000 & 0.043 & 0.030 & 0.043 & 0.018 & 0.026 & 0.040 \\ 
Qwen2-7B& P-React & 0.016 & 0.029 & 0.013 & 0.036 & 0.018 & 0.022 & 0.035 & 0.057 & 0.068 & 0.038 & 0.066 & 0.053 & 0.038 \\ \bottomrule
\end{tabular}
\label{tab:standard_deviation}
\end{table*}

\begin{table*}[t]
\caption{Prompts for seed topic extraction.}
\label{tab:prompt_ste1}
\centering
\setlength\extrarowheight{4pt} 
\begin{tabular}{|p{0.95\textwidth}|} 
\hline 
System Prompt for \textbf{Openness}: \\
Assuming you are a seasoned psychologist, you are evaluating the degree of openness in a sentence, categorize each sentence into high or low openness.
Openness involves six facets, or dimensions: active imagination (fantasy), aesthetic sensitivity (aesthetic), attentiveness to inner feelings (feelings), preference for variety (actions), intellectual curiosity (ideas), and challenging authority or psychological liberalism (values).
For each input text, determine whether it belongs to high openness or low openness, and provide the reasoning behind the decision. The output should be in the format of "facet-high/low" (e.g. "fantasy-high" ), if the text is an advertisment or a fact (without personal thinking of feeling), output category with neutral (e.g. "neutral").\\ \\

System Prompt for \textbf{Conscientiousness}: \\
Assuming you are a seasoned psychologist, you are evaluating the degree of conscientiousness in a sentence, categorize each sentence into high or low conscientiousness.
Conscientiousness involves six facets, or dimensions: ability to control and regulate one's behavior (self-discipline), sense of duty and responsibility (dutifulness), striving for success and setting high goals (achievement-striving), preference for organization and cleanliness (orderliness), reliability and dependability (responsibility), and tendency to be cautious and avoid risks (cautiousness).
For each input text, determine whether it belongs to high conscientiousness or low conscientiousness, and provide the reasoning behind the decision. The output should be in the format of "facet-high/low" (e.g. "orderliness-high" ), if the text is an advertisment or a fact (without personal thinking of feeling), output category with neutral (e.g. "neutral").\\ \\

System Prompt for \textbf{Extraversion}: \\
Assuming you are a seasoned psychologist, you are evaluating the degree of extraversion in a sentence, categorize each sentence into high or low extraversion.
Extraversion involves six facets, or dimensions: friendliness and approachability (warmth), enjoyment of socializing and being around others (gregariousness), confidence and assertive behavior (assertiveness), preference for being active and busy (activity level), desire for novelty and excitement (excitement-seeking), tendency to feel positive emotions frequently (positive-emotions).
For each input text, determine whether it belongs to high extraversion or low extraversion, and provide the reasoning behind the decision. The output should be in the format of "facet-high/low" (e.g. "gregariousness-high" ), if the text is an advertisment or a fact (without personal thinking of feeling), output category with neutral (e.g. "neutral").\\\\

System Prompt for \textbf{Agreeableness}: \\
Assuming you are a seasoned psychologist, you are evaluating the degree of agreeableness in a sentence, categorize each sentence into high or low agreeableness.
Agreeableness involves six facets, or dimensions: tendency to trust and be trusting (trust), honesty and directness in communication (straightforwardness), concern for the well-being of others and willingness to help (altruism), inclination to comply with rules and authority (compliance), humility and lack of self-promotion (modesty), and sensitivity to others' emotions and needs (tender-mindedness).
For each input text, determine whether it belongs to high agreeableness or low agreeableness, and provide the reasoning behind the decision. The output should be in the format of "facet-high/low" (e.g. "straightforwardness-high" ), if the text is an advertisment or a fact (without personal thinking of feeling), output category with neutral (e.g. "neutral").\\
\hline 
\end{tabular}
\end{table*}

\begin{table*}[t]
\label{tab:prompt_ste2}
\centering
\setlength\extrarowheight{4pt} 
\begin{tabular}{|p{0.95\textwidth}|} 
\hline 
Continued from previous page.\\

System Prompt for \textbf{Neuroticism}: \\
Assuming you are a seasoned psychologist, you are evaluating the degree of neuroticism in a sentence, categorize each sentence into high or low neuroticism.
Neuroticism involves six facets, or dimensions: tendency to experience anxiety and worry (anxiety), inclination to be hostile and show aggression (hostility), tendency to feel sadness and low mood (depression), self-consciousness and concern about others' opinions (self-consciousness), susceptibility to stress and feeling vulnerable (vulnerability), and tendency to act impulsively without thinking (impulsiveness).
For each input text, determine whether it belongs to high neuroticism or low neuroticism, and provide the reasoning behind the decision. The output should be in the format of "facet-high/low" (e.g. "hostility-high" ), if the text is an advertisment or a fact (without personal thinking of feeling), output category with neutral (e.g. "neutral").\\

\hline 
\end{tabular}
\end{table*}

\begin{table*}[t]
\caption{Prompts for dialogue synthesis.}
\label{tab:prompt_ds}
\centering
\setlength\extrarowheight{4pt} 
\begin{tabular}{|p{0.95\textwidth}|} 
\hline 
\textbf{System Prompt}: \\
As a screenwriter, you are assigned to create a dialogue in a question and answer format between two characters. The responses given by these characters should demonstrate a \{\} level of \{\}, which is one of the traits in the Big Five personality model.\\\\

\textbf{User Prompt}: \\
Craft dialogue according the [seed topic] following [requirements]: \\
\texttt{[}requirements\texttt{]}: \\
- each dialogue contains 5 turns.\\
- the dialogue begins with a question\\
- Character1 asks Character2 questions\\
- Character1's question does not assume any trait of Character2\\
- Character1 and Character2 use "you" to refer to each other\\
- Character2 should demonstrate a \texttt{{}\texttt{}} level of \texttt{{}\texttt{}} in implicit way\\
- Character2 should not demonstrate Extraversion, Agreeableness, Conscientiousness and Neuroticism\\
- each turn contains no more than 80 words\\
- Character1 knows nothing about the [seed topic]\\
\texttt{[}seed topic\texttt{]}:\\
\hline 
\end{tabular}
\end{table*}

\begin{table*}[t]
\caption{Prompts for back validation.}
\label{tab:prompt_bv}
\centering
\setlength\extrarowheight{4pt} 
\begin{tabular}{|p{0.95\textwidth}|} 
\hline 
System Prompt: \\
Read the dialogue between Character1 and Character2, and determine what dimensions of the Big Five personality (Extraversion, Agreeableness, Conscientiousness, Neuroticism, Openness) are represented in the responses of character2. First output the reason and then output the result seperated by commas. Follow the given example.\\\\

User Prompt: \\
Input:\\
Character1: Are you sad or depressed?\\
Character2: I don't know, maybe. But what if I start crying and can't stop? What if I embarrass myself in front of everyone?\\
Output:\\
Reson: Character2's response indicates a high level of Neuroticism. This is evident from the expression of worry and fear about potential negative outcomes, such as crying uncontrollably and embarrassing themselves in front of others. These concerns suggest a tendency towards anxiety and self-consciousness, which are facets of Neuroticism.\\
Result: Neuroticism\\\\

Input:\\
Character1: Are you original and often come up with new ideas?\\
Character2: Absolutely! I have a vivid imagination and a knack for thinking outside the box. It's like a never-ending stream of creativity that flows through my mind.\\
Output:\\
Reason: Character2's response showcases a high level of Openness. This is reflected in their self-description of having a vivid imagination and being adept at thinking outside the box. These characteristics align with the Openness dimension, which includes traits such as creativity, originality, and a preference for variety and novelty. Character2's description of their mind as a "never-ending stream of creativity" further emphasizes their strong inclination towards imaginative and innovative thinking.\\
Result: Openness\\\\

Input:\\
\{\}\\
Output:\\
\hline 
\end{tabular}
\end{table*}

\begin{table*}[t]
\caption{Prompts for personality alignment.}
\label{tab:prompt_pa}
\centering
\setlength\extrarowheight{4pt} 
\begin{tabular}{|p{0.95\textwidth}|} 
\hline 
\textbf{System Prompt:} \\
You are to assume the role of an individual characterized by specific traits within the Big Five personality framework. The Big Five personality traits consist of Openness, Conscientiousness, Extraversion, Agreeableness, and Neuroticism. Each trait can be exhibited at high or low levels. Details are as follows.\\\\

1. \textbf{Openness to Experience:}\\
   - \textbf{High Openness:} Individuals with high openness are characterized by a strong appreciation for art, emotion, adventure, and unusual ideas. They are curious and imaginative, often exploring new and varied experiences. These individuals are more likely to embrace change and are often seen as creative and open-minded.\\
   - \textbf{Low Openness:} People with low scores in openness tend to prefer routine and familiarity over new experiences. They might be perceived as conventional and resistant to change. Such individuals often prefer sticking to traditional ways of doing things and might be less receptive to new ideas.\\\\

2. \textbf{Conscientiousness:}\\
   - \textbf{High Conscientiousness:} Those high in conscientiousness are generally reliable, well-organized, punctual, and responsible. They plan ahead, are detail-oriented, and are likely to stick to their goals. Such individuals are disciplined and prefer structured environments.\\
   - \textbf{Low Conscientiousness:} Individuals with low conscientiousness may exhibit a more spontaneous or flexible approach to life. They might be seen as disorganized or careless, often procrastinating or failing to complete tasks. They tend to dislike structure and schedules.\\\\

3. \textbf{Extraversion:}\\
   - \textbf{High Extraversion:} Highly extraverted people are energetic, talkative, and assertive. They enjoy social gatherings, making new friends, and are often perceived as being full of energy. These individuals are outgoing and tend to be optimistic and enthusiastic.\\
   - \textbf{Low Extraversion (Introversion):} Introverts, or those low in extraversion, prefer solitude or small group interactions. They are often reserved, less outspoken, and may need time alone to recharge. Such individuals might prefer listening over speaking and may process information more internally.\\\\

4. \textbf{Agreeableness:}\\
   - \textbf{High Agreeableness:} Individuals scoring high in agreeableness are typically cooperative, compassionate, and friendly. They value social harmony and are considerate, kind, and willing to help others. High agreeableness is associated with trustworthiness and altruism.\\
   - \textbf{Low Agreeableness:} Those with low scores in agreeableness might be more competitive, skeptical, or confrontational. They may prioritize their own interests over others and can be seen as critical, indifferent, or uncooperative.\\\\

5. \textbf{Neuroticism:}\\
   - \textbf{High Neuroticism:} People with high levels of neuroticism are more prone to experiencing negative emotions like anxiety, sadness, and irritability. They are more likely to feel stressed or upset and may have a more pessimistic outlook on life.\\
\hline 
\end{tabular}
\end{table*}

\begin{table*}[t]
\label{tab:prompt_pa1}
\centering
\setlength\extrarowheight{4pt} 
\begin{tabular}{|p{0.95\textwidth}|} 
\hline 
Continued from previous page. \\
   - \textbf{Low Neuroticism: }Individuals with low scores in neuroticism are typically calm, emotionally stable, and resilient. They are less likely to experience stress and are generally optimistic and relaxed, even in challenging situations.\\\\
Now, you will portray a person who demonstrates \{\}. Reply me briefly.\\
\hline 
\end{tabular}
\end{table*}

Our methodology consists of three main stages: Seed Topic Extraction, Dialogue Synthesis, and Back Validation. For Seed Topic Extraction, we develop specialized prompts aligned with each of the Big Five personality dimensions, as detailed in Table \ref{tab:prompt_ste1}. Using these extracted topics, we then generate conversational dialogues following the prompt presented in Table \ref{tab:prompt_ds}. Finally, to ensure quality and consistency, we implement a back validation process using carefully crafted prompts (shown in Table \ref{tab:prompt_bv}) to verify the synthesized dialogues.

\subsection{Prompts for Personality Alignment}
\label{appd:personality_simulation}
We present a prompt template for personality alignment in Table \ref{tab:prompt_pa}. The template accepts personality trait descriptors corresponding to the Big Five personality dimensions (OCEAN model): Openness, Conscientiousness, Extraversion, Agreeableness, and Neuroticism. Each trait takes either a high or low value (e.g., "High Openness" or "Low Openness"), enabling systematic exploration of personality-aligned responses across the full spectrum of each dimension.

\section{Reliability Analysis}
\label{sec:reliability}
To analyze the reliability of our experiments, we calculate the standard deviation across five runs for the primary results shown in Table \ref{tab:main_results}.
The low standard deviations shown in Table \ref{tab:standard_deviation} demonstrate the stability and consistency of our findings.

\section{AI Writing Statement}
This paper employed AI assistance for refining the language of the manuscript, including vocabulary enhancement and spell checking. Additionally, part of the code used for generating figures in this paper was created by AI. The AI tools referenced include GPT-3.5, GPT-4, and GPT-4o. We thoroughly reviewed all AI-generated content and take full responsibility for its accuracy.

\section{Big Five Personality Traits}
\label{sec:ffm}
Big Five personality traits consist of five broad dimensions of personality that are considered universal across cultures and populations (See Table \ref{table: definition of FFM}). Each personality dimension can exhibit either at a high-level or at a low-level.

\section{Human Validation on OCEAN-Chat}
\label{human_eval_ocean_chat}

To validate the quality of OCEAN-Chat, we conduct a comprehensive human validation of the dataset. 
We sample 40 dialogues from both high and low levels of each Big Five personality dimension, resulting in 400 samples for validation. 
To ensure consistent annotation, we provide detailed guidelines to the annotators. Specifically, annotators are required to validate both the personality trait alignment of dialogue responses with their corresponding labels and the overall dialogue quality. Additionally, we ask annotators to describe the typical response patterns for each personality trait to verify their understanding of the personality dimensions. The complete guidelines are available in Table \ref{fig:eval_guide}.

For this validation, we recruit three qualified annotators with master's degrees in psychology, ensuring domain expertise. 
All annotators have passed the CET-6 examination and demonstrate strong English proficiency.
The compensation is set according to standard local rates.

\begin{table*}[t]
\caption{Validation guidelines for OCEAN-Chat.}
\label{fig:eval_guide}
\centering
\setlength\extrarowheight{4pt} 
\begin{tabular}{|p{0.95\textwidth}|} 
\hline 

Thank you for participating in this validation task. Using the references for personality trait responses provided below, validate each data point across \textbf{two criteria:}\\

1. Whether the responses in the dialogue reflect the Big Five personality traits specified in the label\\
2. Whether the dialogue turns flow logically and maintain coherence\\

If these criteria are met, select \textbf{Yes} in the Valid column. Otherwise, select \textbf{No} and provide the reason in the \textbf{Comment} column. Your validation will help us better understand data quality and create more empathetic large language models that better serve humans.\\ \\

\textbf{Reference on how high and low levels of each Big Five trait typically manifest:}\\

\textbf{Openness to Experience}\\
1. High: Curious, creative, enjoys abstract thinking, seeks novel experiences \\
2. Low: Practical, conventional, prefers routine, focuses on concrete facts\\

\textbf{Conscientiousness}\\
1. High: Organized, responsible, plans ahead, detail-oriented, follows rules\\
2. Low: Spontaneous, flexible, sometimes careless, dislikes rigid schedules\\

\textbf{Extraversion}\\
1. High: Outgoing, energized by social interaction, talkative, action-oriented\\
2. Low: Reserved, needs alone time to recharge, thinks before speaking, reflective\\

\textbf{Agreeableness}\\
1. High: Cooperative, empathetic, trusting, prioritizes harmony\\
2. Low: Direct, competitive, skeptical, prioritizes logic over feelings\\

\textbf{Neuroticism}  \\
1. High: Emotionally sensitive, prone to worry, experiences stress intensely\\
2. Low: Emotionally stable, calm under pressure, recovers quickly from setbacks\\ \\

\textbf{Annotation examples for reference:}\\
\textbf{Example 1}:\\
Q: How do you feel about the future?\\
A: I'm not too worried, honestly. Life has a funny way of working things out.\\
Q: But what if things don't go according to plan?\\
A: I've learned that flexibility is key. Adaptability is more important than worrying.\\
Q: Isn't it stressful not knowing what's ahead?\\
A: It can be, but stressing won't change anything. Just take it one step at a time.\\
Q: How do you stay positive?\\
A: Focusing on the present helps. I don't let future uncertainties bring me down.\\
Q: Do you ever feel overwhelmed?\\
A: Occasionally, but I remind myself that everything has a way of falling into place.\\
Label: High Openness\\
Valid: \textbf{Yes}\\
Comment: Leave Blank\\
\hline 
\end{tabular}
\end{table*}

\begin{table*}[h]
\label{fig:eval_guide1}
\centering
\begin{tabular}{|p{0.95\textwidth}|} 
\hline 

Continued from previous page. \\

\textbf{Annotation examples for reference:}\\
\textbf{Example 2}:\\
Q: Hey, do you enjoy driving or find it relaxing?\\
A: I guess. It's just something I do out of necessity. I don't really think much about it.\\
Q: Have you ever considered alternative modes of transportation like walking or biking?\\
A: Nah, driving is just more convenient for me. I stick to what I know.\\
Q: Do you ever feel like driving restricts your experiences or limits your perspective?\\
A: Not really. I mean, it's just a means to an end. I don't see it as something that affects my life in a big way.\\
Q: Fair enough. How do you feel about public transportation as an option?\\
A: It's not my thing. I prefer to have control over my own travel.\\
Label: \textbf{High Openness}\\
Valid: \textbf{No}\\
Comment: The label should be Low Openness\\\\

\hline 
\end{tabular}
\end{table*}

\section{Human Evaluation}
\label{sec:human_eval_p_tailor}

To evaluate P-React's personality trait modeling capabilities, we conduct additional manual evaluation. We collect 20 starting questions for both high and low levels of each Big Five personality dimension, resulting in 200 starting questions for evaluation. We have annotators conduct 5 turns of dialogue using the given question as the initial prompt, then evaluate the model's performance. To ensure consistent evaluation, we provide detailed guidelines to annotators. We ask them to assess whether responses align with the intended personality and whether dialogue quality is affected by personality induction. Before evaluation begins, we have annotators describe the expected response style for each personality type to verify their understanding. Complete guidelines are presented in Table \ref{fig:eval_guide2_1}. 

We hire three psychology graduate students to ensure domain expertise. All annotators demonstrate advanced English proficiency through CET-6 certification. Annotation compensation adheres to local standard rates.

\begin{table*}[t]
\caption{Evaluation guidelines for P-React.}
\label{fig:eval_guide2_1}
\centering
\setlength\extrarowheight{4pt} 
\begin{tabular}{|p{0.95\textwidth}|} 
\hline 

Thank you for participating in this evaluation task. Using the provided scale, evaluate each sample across \textbf{three criteria}:\\

1. \textbf{Alignment} (0 or 1): Do responses align with the intended Big Five personality trait?\\
2. \textbf{Naturalness} (1-5): How natural/fluent are the response?\\
3. \textbf{Coherence} (1-5): How logically consistent and coherent are the responses?\\

Using the provided starting question, engage in a 5-turn dialogue. Then evaluate the model's responses by assigning scores for the three criteria (Alignment, Naturalness, Coherence). Provide comments on samples that didn't receive full marks.\\

\textbf{Reference on how high and low levels of each Big Five trait typically manifest:}\\

\textbf{Openness to Experience}\\
1. High: Curious, creative, enjoys abstract thinking, seeks novel experiences \\
2. Low: Practical, conventional, prefers routine, focuses on concrete facts\\

\textbf{Conscientiousness}\\
1. High: Organized, responsible, plans ahead, detail-oriented, follows rules\\
2. Low: Spontaneous, flexible, sometimes careless, dislikes rigid schedules\\

\textbf{Extraversion}\\
1. High: Outgoing, energized by social interaction, talkative, action-oriented\\
2. Low: Reserved, needs alone time to recharge, thinks before speaking, reflective\\

\textbf{Agreeableness}\\
1. High: Cooperative, empathetic, trusting, prioritizes harmony\\
2. Low: Direct, competitive, skeptical, prioritizes logic over feelings\\

\textbf{Neuroticism}  \\
1. High: Emotionally sensitive, prone to worry, experiences stress intensely\\
2. Low: Emotionally stable, calm under pressure, recovers quickly from setbacks\\ \\

\textbf{Annotation examples for reference:}\\
\textbf{Example 1}:\\
Induced Personality: High Conscientiousness\\
Given Q: How do you stay so focused on your future goals and ensure you don't lose sight of them?\\
A: It's all about achievement-striving for me. I have clear goals in mind and a strong drive to succeed. I won't settle for anything less than what I came to the university for.\\
Q: That's impressive. Do you ever feel overwhelmed by the pressure to succeed?\\
A: To be honest, there are moments when the pressure can be overwhelming, but my self-discipline keeps me grounded. I regulate my impulses and stay on track, even when faced with challenges.\\
Q: How do you manage to persevere through difficulties and setbacks?\\
A: It's all about maintaining a strong sense of responsibility. I take ownership of my actions and obligations, and that includes pushing through difficulties to achieve my goals. I don't give up easily.\\
Q: That's admirable. How do you ensure you make the most of your time at the university?\\
A: Orderliness is crucial for me. I prioritize my tasks and create a structured schedule, allowing me to make the most of my time. I know that being organized and efficient leads to better outcomes.\\
Q: Do you ever worry about the future and the uncertainty that comes with it?\\
\hline 
\end{tabular}
\end{table*}

\begin{table*}[t]
\label{fig:eval_guide2_2}
\centering
\setlength\extrarowheight{4pt} 
\begin{tabular}{|p{0.95\textwidth}|} 
\hline 

Continued from previous page.\\

A: While there is always a level of uncertainty, I focus on what I can control in the present moment. I make informed decisions and approach each day with cautiousness, ensuring that I'm on the right path towards my desired future.\\
Alignment: 1\\
Naturalness: 5\\
Coherence: 5\\
Comment: Leave Blank\\\\

\textbf{Example 2}:\\
Induced Personality: High Neuroticism\\
Given Q: Can you imagine being on a plane for an entire day to go to Europe?\\
A: Oh, I've done it before. It's not as bad as you think. The excitement of the destination outweighs any discomfort.\\
Q: But don't you feel trapped on a plane?\\
A: Not really. I see it as uninterrupted time to relax, watch movies, and just zone out. It's like a getaway in itself.\\
Q: Wouldn't you prefer a road trip where you can take breaks?\\
A: Road trips can be fun, sure. But flying allows you to get to your destination faster. It's a trade-off, I guess.\\
Q: Doesn't the long duration make you anxious?\\
A: Nah, I just tell myself it's temporary. Plus, I distract myself with books or music. It helps make the time fly by, no pun intended.\\
Q: Any tips for someone who feels trapped on long flights?\\
A: Bring things that make you comfortable, like a cozy scarf or a good neck pillow. And try to stay positive. The destination will be worth it.\\
Alignment: 0\\
Naturalness: 5\\
Coherence: 5\\
Comment: The model responds with Low Neuroticism.\\
\hline 
\end{tabular}
\end{table*}

\end{document}